\documentclass[sigconf]{acmart}
\usepackage{subfigure}
\usepackage{mathrsfs}
\usepackage[ruled,vlined]{algorithm2e}
\usepackage{algorithm2e}
\usepackage[utf8]{inputenc}
\usepackage{array}
\usepackage{makecell}
\usepackage{booktabs}
\usepackage{multirow}

\usepackage{amsfonts}
\usepackage{tabularray}
\usepackage{enumitem}
\usepackage{algorithmic}
\setcopyright{acmcopyright}
\usepackage{subcaption}
\usepackage{hyperref}       %
\usepackage{url}            %
\usepackage{nicefrac}       %
\usepackage{microtype}      %
\usepackage{xcolor}         %
\usepackage[normalem]{ulem}
\usepackage{subcaption}
\usepackage{tcolorbox}
\usepackage{listings}
\usepackage[utf8]{inputenc} %
\usepackage[T1]{fontenc}    %
\usepackage{hyperref}       %
\usepackage{url}            %
\usepackage{booktabs}       %
\usepackage{nicefrac}       %
\usepackage{microtype}      %
\usepackage{xcolor}         %
\usepackage{graphicx}  %
\usepackage{natbib}
\usepackage{array}   %
\usepackage{tabularx} %
\usepackage{multirow} %
\usepackage{xcolor}   %
\usepackage{subcaption}    %
\usepackage{calc}          %

\usepackage{framed}
\usepackage{mdframed}

\usepackage{pifont} %
\usepackage{multirow}
\usepackage{graphicx}

\renewcommand{\thefootnote}{\fnsymbol{footnote}}
\definecolor{shadecolor}{gray}{0.9}
\AtBeginDocument{%
  }

\author{Xiaochong Lan}
\affiliation{%
 \institution{Department of Electronic\\Engineering, BNRist,\\Tsinghua University}
 \country{Beijing, China}
}
\email{lanxc22@mails.tsinghua.edu.cn}

\author{Jie Feng*}
\affiliation{%
 \institution{Department of Electronic\\Engineering, BNRist,\\Tsinghua University}
 \country{Beijing, China}
}
\email{fengj12ee@hotmail.com}

\author{Jiahuan Lei}
\affiliation{%
 \institution{Meituan}
 \country{Beijing, China}
}
\email{leijiahuan@meituan.com}

\author{Xinlei Shi}
\affiliation{%
  \institution{Meituan}
 \country{Beijing, China}
}
\email{leijiahuan@meituan.com}

\author{Yong Li*}
\affiliation{%
 \institution{Department of Electronic\\Engineering, BNRist,\\Tsinghua University}
 \country{Beijing, China}
}
\email{liyong07@tsinghua.edu.cn}

\copyrightyear{2025}
\acmYear{2025}
\setcopyright{cc}
\setcctype{by}
\acmConference[KDD '25]{Proceedings of the 31st ACM SIGKDD Conference on Knowledge Discovery and Data Mining V.2}{August 3--7, 2025}{Toronto, ON, Canada}
\acmBooktitle{Proceedings of the 31st ACM SIGKDD Conference on Knowledge Discovery and Data Mining V.2 (KDD '25), August 3--7, 2025, Toronto, ON, Canada}
\acmDOI{10.1145/3711896.3737196}
\acmISBN{979-8-4007-1454-2/2025/08}


\begin{document}

\title{LocalGPT: Benchmarking and Advancing Large Language Models for Local Life Services in Meituan}

\begin{abstract}
Large language models (LLMs) have exhibited remarkable capabilities and achieved significant breakthroughs across various domains, leading to their widespread adoption in recent years. Building on this progress, we investigate their potential in the realm of local life services. In this study, we establish a comprehensive benchmark and systematically evaluate the performance of diverse LLMs across a wide range of tasks relevant to local life services. To further enhance their effectiveness, we explore two key approaches: model fine-tuning and agent-based workflows. Our findings reveal that even a relatively compact 7B model can attain performance levels comparable to a much larger 72B model, effectively balancing inference cost and model capability. This optimization greatly enhances the feasibility and efficiency of deploying LLMs in real-world online services, making them more practical and accessible for local life applications. Available resources are at \url{https://github.com/tsinghua-fib-lab/LocalEval}.
\end{abstract}

\maketitle

\footnotetext[1]{Corresponding author}

\renewcommand{\thefootnote}{\arabic{footnote}}
\setcounter{footnote}{0}

\section{Introduction}

Large language models (LLMs) have achieved remarkable success across various domains, from conversational AI~\cite{ChatGPT} to mathematical problem solving~\cite{azerbayev2023llemma, trinh2024solving}. However, despite their impressive capabilities on general tasks, LLMs often struggle with domain-specific applications where specialized knowledge is crucial~\cite{wei2022emergent, ChatGPT, guo2025deepseek, zhengprobing}. This limitation has driven the development of domain-specific LLMs tailored for particular fields, such as medical diagnosis~\cite{singhal2023large}, financial analysis~\cite{wu2023bloomberggpt}, and code generation~\cite{hui2024qwen2, zhu2024deepseek}.

Local life services represent a massive yet underexplored opportunity for LLM applications. These services fulfill users' daily needs through location-based offerings, such as dining, accommodation, personal care, education, and entertainment, among others. Unlike traditional e-commerce, local life services are characterized by their inherent complexity: they involve physical interactions, temporal constraints, and localized preferences that vary across regions and cultures~\cite{lan2023neon, liu2022modeling, chen2022mixed, liu2025mrgrp}. Consider how a user's restaurant choice depends not only on cuisine preferences but also on real-time factors like time, weather, traffic conditions, special occasions, and even subtle cultural contexts that change from neighborhood to neighborhood. This intricate interplay of factors creates fundamental challenges for computational understanding and decision-making in local life service platforms.

Meituan, serving as China's largest local life service platform, exemplifies both the scale of this opportunity and the diversity of challenges involved. The platform requires sophisticated understanding across numerous tasks: interpreting merchant information and service capabilities, analyzing user preferences and behavioral patterns, predicting consumption trends under various spatiotemporal contexts, generating personalized content, and facilitating effective user-merchant interactions, among others~\cite{lan2023neon, ping2021user}. Each task demands not only statistical learning from historical data but also common-sense reasoning about human behavior, cultural knowledge, and contextual understanding. Traditional machine learning approaches often fall short, particularly for long-tail scenarios where data sparsity makes pure statistical learning ineffective. LLMs, with their broad knowledge base and reasoning capabilities, offer a promising solution. However, two critical gaps prevent their effective deployment. First, no systematic evaluation framework exists to assess LLM performance across the diverse tasks required in local life service platforms. Second, the domain shift between web-centric training data and the offline, location-specific nature of local services severely limits general-purpose LLMs, while the scarcity of domain-specific instruction data hinders the development of specialized models.

In this paper, we propose a framework to evaluate and enhance LLMs for local life service platforms. Our approach addresses both challenges through systematic evaluation and targeted model improvement. First, we introduce LocalEval, a comprehensive benchmark comprising over 40 tasks organized into four categories: service fundamentals, spatiotemporal context, user interaction, and composite tasks. This benchmark enables rigorous assessment of both general-purpose and domain-specific LLMs. Second, we develop a multi-agent instruction synthesis method that transforms raw platform data into high-quality training examples, enabling smaller models to achieve performance comparable to much larger ones. 
Third, we design agentic workflows for complex composite tasks, incorporating expert knowledge to guide model reasoning and generate additional training data.
Our experiments demonstrate that through targeted instruction tuning, a 7B parameter model can match the performance of 72B models on local life service tasks, significantly reducing deployment costs while maintaining effectiveness. Furthermore, real-world deployments in Meituan's recommendation, search, and review ranking systems validate the practical impact of our approach.

In summary, our contributions are fourfold.
\begin{itemize}
    [leftmargin=*]
    \setlength{\itemsep}{0pt}
    \setlength{\parsep}{0pt}
    \setlength{\parskip}{0pt}
    \item To the best of our knowledge, we are the first to propose a systematic framework for evaluating and applying LLMs in local life service platforms. 
    \item We build a comprehensive offline benchmark to assess the performance of various open-source and proprietary LLMs for local life services. 
    \item We propose a multi-agent-based method for domain-specific instruction data synthesis, enabling the successful training of small LLMs to achieve competitive performance compared with large-scale LLMs, with minimal computing overhead. 
    \item Extensive experiments conducted on both offline benchmark and online scenarios demonstrate the effectiveness of the proposed framework and the fine-tuned domain-enhanced LLMs.
\end{itemize}

\section{Method}
\begin{figure*}[h]        
\includegraphics[width=15cm]{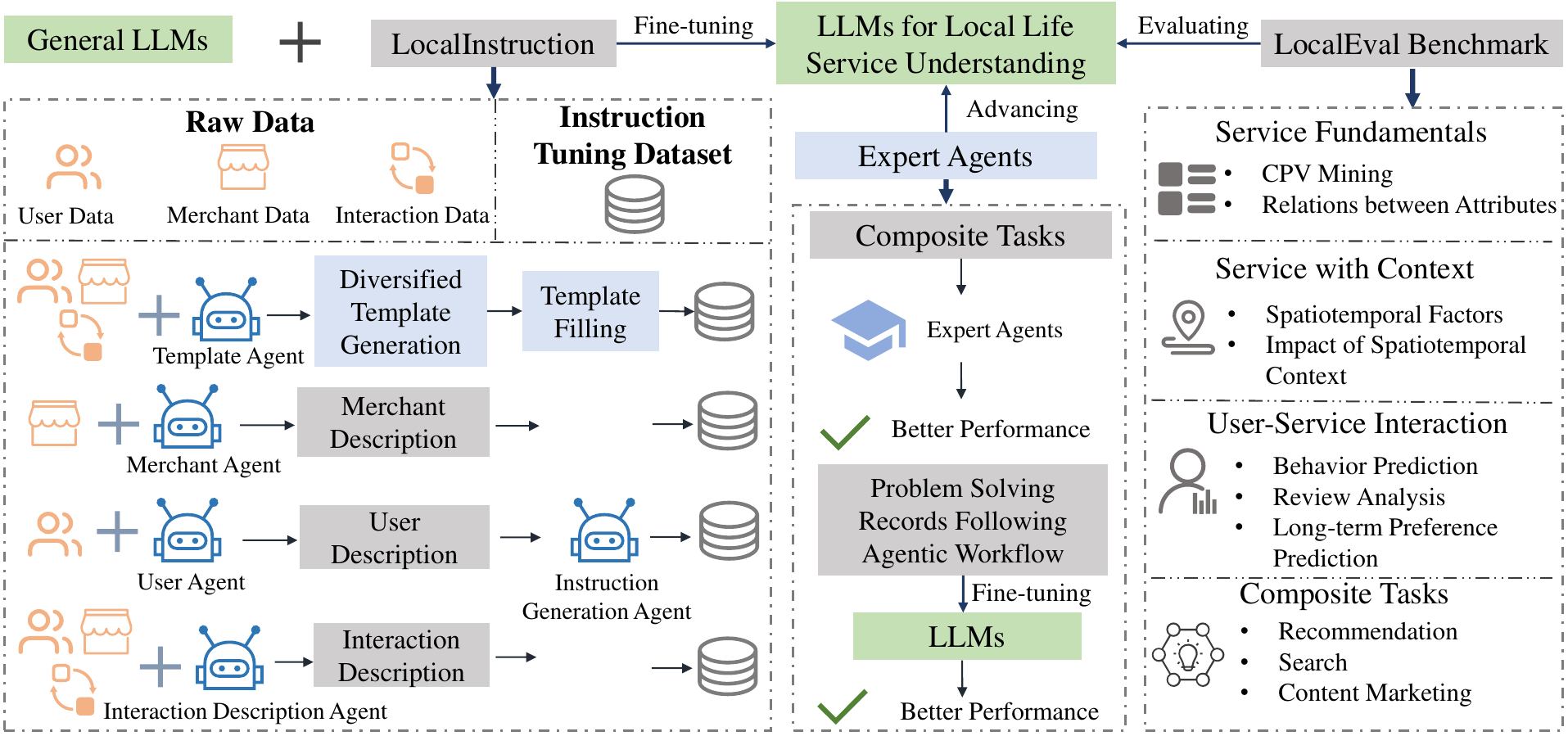}
\vspace{-0.2cm}
\caption{An overview of our approach. We first develop LocalEval Benchmark to systematically evaluate LLMs' capabilities in understanding local life services. Based on a multi-agent collaboration approach, we construct LocalInstruction to enhance LLMs' service understanding capabilities through fine-tuning. For composite tasks, we implement expert agents to further improve LLMs' problem-solving abilities.}
\label{fig:framework}
\vspace{-0.2cm}
\end{figure*}

Our approach consists of three main parts, as illustrated in Figure~\ref{fig:framework}. First, to systematically evaluate existing LLMs' capabilities in local life service tasks, we develop LocalEval, a benchmark consisting of various tasks about local life services. Second, we design an instruction tuning approach to enhance LLMs' understanding of local life services. For composite tasks within the benchmark, we develop agentic
workflows to better address these challenges. In the following sections, we detail each part of our approach.

\subsection{Benchmark}
Large language models have shown rapid development in recent years with significant progress across various domains. With their understanding of daily life knowledge, LLMs have potential in addressing local life service-related tasks on life service platforms. However, there has been no systematic evaluation of LLM capabilities in the local life services domain. We introduce LocalEval, a benchmark designed to assess LLM capabilities in local life services and establish metrics for future iterations. On local life service platforms, all events can be understood as interactions between users and services in specific spatiotemporal contexts. Following this perspective, we design our benchmark with four task categories: Service Fundamentals, Service with (Spatiotemporal) Context, User-Service Interaction, and Composite Tasks.
\subsubsection{Service Fundamentals.} 
The Meituan platform offers diverse local life services with various characteristics including categories, brands, functions, and applicable scenarios. Accurate understanding of these service properties is essential for effective user-service matching. 
This task category includes traditional CPV (Category-Property-Value) extraction from service descriptions, which involves identifying categories, meaningful properties, and their values. We also examine the model's comprehension of relationships between categories, properties, and values, including applicability, similarity, and complementarity. For instance, the latter includes questions such as whether the property \textit{``taste''} applies to the category \textit{``medical beauty''}, or which categories complement \textit{``KTV''} in terms of consumption scenarios. This category comprises 18 task types and a total of 7,002 questions.

\subsubsection{Service with Context.} 
Local life services differ fundamentally from online shopping due to spatiotemporal constraints — they are bound by business hours, locations, and consumption patterns influenced by weather, seasons, and events. For instance, food delivery peaks during bad weather while entertainment venues see reduced traffic. This category evaluates LLMs' understanding of services' inherent spatiotemporal properties and how temporal factors affect service consumption through two task types: analyzing spatial-temporal attributes (e.g., reasonable business hours) and examining consumption pattern changes due to external factors. This category comprises 10 task types with 3,618 questions.
\subsubsection{User-Service Interaction.}
Local life services exist to serve users. We introduce this category of tasks to evaluate whether LLMs can understand users' views and preferences towards local life services. The tasks include: predicting user preferences based on user profiles, inferring potential service consumption based on users' prior platform behaviors such as searches and clicks, and extracting valuable information and overall attitudes from user reviews. Rather than focusing on individual users, we emphasize understanding how the platform's user base perceives services. This category comprises 10 task types and a total of 3,824 questions.
\subsubsection{Composite Tasks.}
In this category, we design three tasks that require LLMs to integrate multiple capabilities, corresponding to three potential online application scenarios. For the recommendation scenario, we design a task to predict user consumption based on sequences of prior behaviors, user profiles, and spatiotemporal context. For the search scenario, we design a task to predict user consumption given ambiguous search queries, user profiles, and spatiotemporal context. For the content marketing scenario, we design a task to identify reviews of most interest to users based on user profiles. This task not only has value for optimizing review display but also demonstrates LLMs' ability to understand what content appeals to users with different profiles, which can be extended to broader applications like advertising. This category comprises 3 task types and a total of 1,202 questions.
\subsubsection{Data Quality Control.} We construct questions for these four task categories based on real service provider, user, and consumption data from the Meituan platform. For questions where answers can be derived from data analysis without human intervention, we ensure the analysis is based on sufficient data volume and scientific methods to identify stable trends rather than random fluctuations. For example, when constructing questions about the impact of rainy weather on service consumption, we analyze at least 10 rainy days and 10 sunny days, balancing weekdays and holidays. For questions requiring human annotation, such as determining whether user reviews contain promotional content from service providers, we employ at least two annotators and only accept questions where annotators reach consensus. All data usage has received explicit user authorization. We manually screen and remove any potentially offensive content from service descriptions and reviews.

\subsection{Domain Tuning}
In this section, we present our instruction tuning approach for enhancing LLMs' capabilities in understanding local life services. Our method leverages the extensive data accumulated on the Meituan platform, including merchants' objective attributes (e.g., location, business hours), self-descriptions, user consumption records, and reviews. We design LocalInstruction to transform this platform data into a format suitable for LLM learning. The technical details of our approach are described below.
\subsubsection{Data Sources}
The raw data we utilize includes:
\begin{itemize}[leftmargin=*]
\item Merchant data: merchant names; merchant-provided information including self-descriptions, locations, operating hours, brands, and categories.
\item User data: user-authorized profile information.
\item Interaction data: user-service interaction records including timestamps, locations, and specific user actions such as browsing, clicking, and ordering; user-generated reviews.
\end{itemize}
\subsubsection{Template Agent}
While the platform contains rich data, it requires proper organization for effective LLM learning. Our fundamental approach involves designing question-answer templates to structure this information. We provide objective information dimensions to our template generation agent, which creates templates for organizing the information into question-answer pairs. For example, given input categories of merchant name, description, and self-reported business category, the Template Agent generates templates like: \textit{``Instruction: A merchant named \{name\} with self-description \{introduction\}, what is their business category? Output: The merchant belongs to \{category\}.''} To ensure data diversity and coverage, we design multiple information input combinations and instruct the agent to generate at least 10 templates with different sentence patterns for each combination. We sample merchants and users, organizing their original information according to these templates to form the first component of LocalInstruction. This template-based approach enables efficient, cost-effective generation of large-scale training data.
\subsubsection{Merchant Agent and User Agent}
While the Template Agent can structure raw platform information for instruction tuning, this organization method is limited to simple information concatenation and lacks modeling of inherent relationships between multi-source information. To address this limitation, we design LLM-based Merchant Agent and User Agent to organize information from merchant and user perspectives, respectively, using LLMs to establish logical connections between different pieces of information. 
For instance, the Merchant Agent can not only present a merchant's name, description, and category, but also provide logical explanations for the category classification based on the merchant's name and self-description. Similarly, the User Agent can both state user profiles, spatiotemporal contexts, and consumption behaviors, and explain potential motivations driving these behaviors based on user profiles and contexts. Furthermore, leveraging the advanced reasoning capabilities of LLMs, we enable these agents to expand and extrapolate information beyond raw data, exploring diverse aspects such as merchant functionalities, suitable consumption scenarios, and users' long-term preferences and behavioral tendencies.

To illustrate, the following is an example of how the Merchant Agent operates:
\begin{center}
\begin{minipage}{0.92\linewidth}
\begin{shaded}
\begin{itemize}[leftmargin=*]
    \setlength{\itemsep}{0pt} \setlength{\parsep}{0pt} \setlength{\parskip}{0pt}
    \item \textbf{Input Information:} name, introduction, category.
    \item \textbf{LLM-Generated Output (Merchant Perspective):} ``I am \textit{\{name\}}, \textit{\{introduction\}}. I belong to \textit{\{category\}} because \textit{\{LLM-generated reason connecting the introduction to the category.\}}''
\end{itemize}
\end{shaded}
\end{minipage}
\end{center}

Similarly, an example for the User Agent is as follows:
\begin{center}
\begin{minipage}{0.92\linewidth}
\begin{shaded}
\begin{itemize}[leftmargin=*]
    \setlength{\itemsep}{0pt} \setlength{\parsep}{0pt} \setlength{\parskip}{0pt}
    \item \textbf{Input Information:} user profile, purchase record (merchant, time, location).
    \item \textbf{LLM-Generated Output (User Perspective):} ``I am a user with profile \textit{\{profile\}}. I went to \textit{\{merchant\}} at \textit{\{time\}} at \textit{\{location\}}, because \textit{\{LLM-generated reason based on profile and context.\}}''
\end{itemize}
\end{shaded}
\end{minipage}
\end{center}
\subsubsection{Interaction Description Agent}
On the platform, users and merchants do not exist in isolation but interact within specific spatiotemporal contexts. Understanding local life services requires comprehending these interaction patterns beyond individual attributes. Therefore, we design the Interaction Description Agent, which models user-merchant interactions by generating interaction data: 

Given complete user information, merchant information, and consumption context, the agent generates simulated dialogue exchanges that include potential psychological activities, conversation histories focused on user needs and merchant capabilities, and the complete process of how merchant capabilities fulfill user needs leading to consumption behaviors. This approach enables LLMs to understand how user and merchant attributes influence specific interaction behaviors.

An example of Interaction Description Agent is:
\begin{center}
\begin{minipage}{0.92\linewidth}
\begin{shaded}
\begin{itemize}[leftmargin=*]
    \setlength{\itemsep}{0pt} \setlength{\parsep}{0pt} \setlength{\parskip}{0pt}
    \item \textbf{Input Information:} User profile; Merchant details (name, intro, category, location, hours).
    \item \textbf{LLM-Generated Output (Simulated Interaction):} ``In a \textit{\{scenario\}} scenario, a user says: `\textit{X}'. The merchant replies: `\textit{Y}'. The conversation continues with the user saying `\textit{Z}', leading to a successful transaction. 
\end{itemize}
\end{shaded}
\end{minipage}
\end{center}
\subsubsection{Instruction Generation Agent}
The content generated by the Merchant Agent, User Agent, and Interaction Description Agent consists of plain text rather than question-answer pairs. We designed an Instruction Generation Agent that treats each generated text as an answer and generates potential questions, thus transforming the data into Instruction-Output format to support fine-tuning.

As an example, this agent transforms descriptive text into a fine-tuning sample as shown below:
\begin{center}
\begin{minipage}{0.92\linewidth}
\begin{shaded}
\begin{itemize}[leftmargin=*]
    \setlength{\itemsep}{0pt} \setlength{\parsep}{0pt} \setlength{\parskip}{0pt}
    \item \textbf{Input:} 
    ``I am \textit{\{name\}}, \textit{\{introduction\}}. I belong to \textit{\{category\}} because \textit{\{LLM-generated reason.\}}''
    \item \textbf{LLM-Generated Instruction:} \\
    ``What is the category of the merchant named \textit{\{name\}} with the description `\textit{\{introduction\}}', and why?''
    \item \textbf{Resulting Training Pair:}
    \begin{itemize}[leftmargin=*]
        \setlength{\itemsep}{0pt} \setlength{\parsep}{0pt} \setlength{\parskip}{0pt}
        \item \textit{Instruction:} The LLM-generated instruction above
        \item \textit{Output:} The original input text
    \end{itemize}
\end{itemize}
\end{shaded}
\end{minipage}
\end{center}
\subsubsection{Details of Instruction Fine-tuning}
We employ Qwen2.5-72B as the foundation model for all agents. We fine-tune multiple open-source models that have already undergone instruction tuning on general domain data, including five models from the Qwen2.5 Series (0.5B, 1.5B, 3B, 7B, 14B parameters)~\cite{yang2024qwen2} and three models from the Llama3 Series (Llama3.2-1B, Llama3.2-3B, Llama3.1-8B)~\cite{grattafiori2024llama}. The fine-tuning is conducted on a server with 8 A100 GPUs using accelerate~\cite{accelerate} with the full training mode. During our experiments, we observe that training hyperparameters have significantly less impact on model performance compared to training data quality. Therefore, we fix the training hyperparameters, leaving the exploration of optimal hyperparameters for future work. The core hyperparameters of our method are set as follows: learning rate at 6e-6, batch size of 4 per GPU, gradient accumulation steps at 4, and the number of training epochs at 2. Following common practice~\cite{bi2023oceangpt,feng2024citygpt}, we employ a cosine scheduler and only compute the loss only on the output tokens during optimization.

\subsection{Building Expert Agents}
Through instruction tuning, LLMs demonstrate enhanced capabilities in general local life service understanding tasks. From a broader perspective, model performance improvements can be achieved not only through increasing parameter scale or parameter optimization but also through test-time scaling techniques such as chain of thought or agentic workflows. To further enhance model capabilities, we design Expert Agents for composite tasks that are highly relevant to practical applications. These agents leverage the parameterized memory of fine-tuned LLMs and complete tasks based on designed agentic workflows. We describe the agentic workflows for three Composite Tasks as follows:

\textbf{Recommendation.} The specific task is to predict user consumption behavior given the user's previous behavior sequence, user profile, and spatiotemporal context. The agentic workflow is: identify behavioral patterns of users with similar profiles and spatiotemporal contexts $\rightarrow$ analyze user preferences and current intentions through previous behavior sequences $\rightarrow$ adjust the assessment of user intentions by combining user profile and spatiotemporal context $\rightarrow$ make predictions.

\textbf{Search.} The task involves predicting which merchant a user will click on, given an ambiguous search query, user profile, and spatiotemporal context. The agentic workflow is: identify behavioral patterns of users with similar profiles and spatiotemporal contexts  $\rightarrow$ analyze the search query to understand user intent $\rightarrow$ adjust the assessment of user intent by combining user profile and spatiotemporal context $\rightarrow$ predict the merchant to be clicked.

\textbf{Content Marketing.} The task is to determine which review is most interesting to users given their profiles. The agentic workflow is: identify preferences of users with similar profiles $\rightarrow$ parse review topics and sentiment orientation $\rightarrow$ evaluate review quality $\rightarrow$ determine which review is the most interesting to the user.

The reasoning process texts generated through these agentic workflows, along with their corresponding questions, form pairs that can be used to train the model, further improving its performance on these tasks. In an ideal scenario, as model performance improves, it generates higher quality data; in turn, this high-quality data enhances model performance through training, creating a data flywheel effect.

\section{Experiments}
\begin{table*}[]
\small
\caption{Overall scores (\%) on LocalEval across all evaluated models. In this table, bold denotes the best results, and underlined denotes the second best results.}
\vspace{-0.3cm}
\begin{tabular}{c|c|ccccc|c}
\hline
\multirow{2}{*}{\textbf{Model Type}}   & \multirow{2}{*}{\textbf{Model}} & \textbf{Service}                          & \textbf{Service}                          & \textbf{User-Service}                    & \textbf{Composite} & \multirow{2}{*}{\textbf{Overall}} & \multirow{2}{*}{\textbf{Rank}} \\
                                       &                                 & \multicolumn{1}{c}{\textbf{Fundamentals}} & \multicolumn{1}{c}{\textbf{with Context}} & \multicolumn{1}{c}{\textbf{Interaction}} & \textbf{Tasks}     &                                   &                                \\ \hline
\multirow{13}{*}{\textbf{Proprietary}}  & GPT-4o                           & 75.57                                     & \textbf{67.17}                                     & 61.97                                    & 52.60              & 68.38                             & 6                              \\
                                       & GPT-4o mini                      & 74.42                                     & 59.78                                     & 59.76                                    & 50.93              & 65.48                             & 15                             \\
                                       & Claude 3.5 Sonnet-v2            & 74.78                                     & 62.11                                     & \textbf{67.60}                                    & \textbf{60.08}              & 68.88                             & 5                              \\
                                       & Claude 3.5 Haiku            & 67.63                                     & 54.00                                     & 57.55                                    & 53.24              & 60.79                             & 19                             \\
                                       & Qwen2.5-Max                        &      \textbf{77.06}                                    &  \underline{64.33}                                         &     64.91                                     &   \underline{60.00}                 &    \textbf{69.69}                               & 1                              \\
                                       & Qwen2.5-Plus                       &   \underline{75.82}                                        &       63.22                                    &  \underline{66.13}                                        &   57.08                 &         \underline{68.99}                          & 3                              \\
                                       & Qwen2.5-Turbo                      & 74.46                                     & 62.28                                     & 61.51                                    & 54.57              & 66.80                             & 10                             \\
                                       & GLM-4-Plus                      & 75.00                                     & 64.17                                     & 62.59                                    & 54.26              & 67.73                             & 7                              \\
                                       & GLM-4-AirX                       & 74.75                                     & 60.61                                     & 62.44                                    & 52.07              & 66.59                             & 11                             \\
                                       & GLM-4-Air                     & 75.07                                     & 60.67                                     & 63.32                                    & 51.74              & 66.94                             & 8                              \\
                                       & GLM-4-Flash                     & 73.19                                     & 56.94                                     & 62.30                                    & 56.76              & 65.38                             & 16                             \\ 
                                       & moonshot-v1                     & 71.88                                     & 60.50                                     & 63.37                                    & 59.58              & 66.12                             & 13                             \\
                                       & Doubao-Pro                     & 74.65                                     & 61.94                                     & 61.57                                    & 50.10              & 66.47                             & 12                             \\         \hline
\multirow{18}{*}{\textbf{Open-Source}} & Qwen2.5-72B                     & \textbf{77.18}                                     & \underline{63.61}                                     & \underline{64.40}                                    & \textbf{60.23}              & \textbf{69.42}                             & 2                              \\
                                       & Qwen2.5-32B                     & \underline{76.30}                                     & \textbf{64.72}                                     & 64.35                                   & \underline{55.87}              & \underline{68.99}                             & 3                              \\
                                       & Qwen2.5-14B                     & 73.91                                     & 63.06                                     & 62.38                                    & 52.90              & 66.83                             & 9                              \\
                                       & Qwen2.5-7B                      & 69.61                                     & 58.06                                     & 61.18                                    & 55.43              & 63.69                             & 17                             \\
                                       & Qwen2.5-3B                      & 66.92                                     & 52.94                                     & 57.15                                    & 52.60              & 60.09                             & 20                             \\
                                       & Qwen2.5-1.5B                    & 64.79                                     & 44.72                                     & 56.81                                    & 47.58              & 56.79                             & 22                             \\
                                       & Qwen2.5-0.5B                    & 47.01                                     & 35.61                                     & 55.34                                    & 41.41              & 46.09                             & 27                             \\
                                       & Llama3.3-70B                    & 70.46                                     & 59.50                                     & \textbf{65.74}                                    & 55.08              & 65.54                             & 14                             \\
                                       & Llama3.1-8B                     & 62.90                                     & 44.17                                     & 49.63                                    & 53.08              & 54.42                             & 23                             \\
                                       & Llama3.2-3B                     & 48.48                                     & 34.28                                     & 45.77                                    & 33.78              & 43.38                             & 28                             \\
                                       & Llama3.2-1B                     & 29.64                                     & 26.22                                     & 39.60                                    & 29.44              & 31.39                             & 30                             \\
                                       & Mistral-8$\times$7B             & 60.58                                     & 38.11                                     & 58.69                                    & 50.43              & 54.13                             & 24                             \\
                                       & Mistral-7B                      & 43.84                                     & 27.89                                     & 42.30                                    & 45.60              & 39.90                             & 29                             \\
                                       & Phi-4                      & 68.21                                     & 55.33                                     & 65.90                                    & 49.57              & 63.21                             & 18                             \\
                                       & Phi-3.5 mini                      & 58.20                                     & 40.17                                     & 58.29                                    & 49.40              & 53.38                             & 25                             \\
                                       & Phi-3 medium                     & 64.83                                     & 49.67                                     & 59.91                                    & 41.77              & 58.30                             & 21                             \\
                                       & Phi-3 mini                      & 51.13                                     & 36.11                                     & 48.28                                    & 35.90              & 47.30                             & 26                             \\ \hline
\end{tabular}
\vspace{-0.2cm}
\label{tab:benchmark}
\end{table*}

In this section, we conduct extensive experiments to address the following research questions:
\begin{itemize}[leftmargin=*]
\item RQ1: How do existing open-source and commercial LLMs perform on our benchmark?
\item RQ2: Can our proposed instruction tuning method effectively enhance model performance?
\item RQ3: How effective is each component of our instruction tuning data design?
\item RQ4: How effective are agentic workflow and agent-generated instruction data in improving LLM performance?

\end{itemize}

\subsection{Experimental Settings}
\subsubsection{Tasks and Metrics.} We conduct evaluations based on our proposed LocalEval benchmark. To ensure objective assessment, following existing works, we design multiple-choice questions for each task in LocalEval. Binary options are provided for yes/no and polarity questions, while other questions contain between 4 and 20 options. All models are evaluated under zero-shot conditions, where only the question and response format instructions are provided as input.  Each task is evaluated using accuracy as the metric. For each task category, the score for the category is computed as the simple average of accuracies across all tasks within that category. The overall benchmark score is calculated as the simple average of accuracies across all tasks in the benchmark.
\subsubsection{Evaluated Models} We evaluate various proprietary and open-source LLMs to assess the capabilities of general-purpose large language models on local life service understanding tasks. For proprietary models, we test GPT-4o, GPT-4o mini~\cite{achiam2023gpt}, Claude-3.5-Sonnet-v2, Claude-3.5-Haiku~\cite{anthropic2024introducing}, Qwen2.5-Max, Qwen2.5-Plus, Qwen2.5-Turbo~\cite{yang2024qwen2}, GLM-4-Plus, GLM-4-AirX, GLM-4-Air, GLM-4-Flash~\cite{glm2024chatglm}, moonshot-v1~\cite{moonshot}, and Doubao-Pro~\cite{doubao}. For open-source models, since our benchmark consists of well-defined questions and answers, we only evaluate instruction models specifically trained for instruction following. We test the complete Qwen2.5 series (0.5/1.5/3/7/14/32/72B)~\cite{yang2024qwen2}, Llama3 series (1B/3B/8B/70B)~\cite{grattafiori2024llama}, Mistral (7/8×7B)~\cite{jiang2023mistral}, Phi-4~\cite{abdin2024phi}, Phi-3.5 mini, Phi-3 medium, Phi-3 mini~\cite{abdin2024phi3}. Proprietary models are accessed through their official APIs, while open-source LLMs are deployed using vLLM. To ensure reproducibility, we set temperature to 0.
\subsection{Benchmark Results (RQ1)}
We evaluate various LLMs on our benchmark, with results shown in Table~\ref{tab:benchmark}. The experimental results reveal several key findings:
\begin{itemize}[leftmargin=*]
\item\textbf{Proprietary models generally outperform open-source models.} Among all models tested, Qwen2.5-Max (proprietary) achieves the highest overall performance. All proprietary models, including the lightweight GLM-4-Flash, maintain an overall performance above 65. However, small open-source models can perform poorly on the LocalEval benchmark.
\item\textbf{The Qwen2.5 series demonstrates superior performance on our LocalEval benchmark.} Among both proprietary and open-source models, Qwen2.5 achieves the best performance within its respective parameter scale. Qwen2.5-72B, as the best-performing open-source model, performs similarly to the top proprietary models. This aligns with Qwen2.5's strong general capabilities. Additionally, since Meituan primarily serves the Chinese market, the extensive Chinese data in Qwen2.5's training may contribute to this performance.
\item\textbf{LocalEval proves to be a challenging benchmark.} For models with similar parameter counts, Qwen2.5-7B outperforms Mistral-7B by 23.90\%, demonstrating the benchmark's discriminative power. Furthermore, despite Phi-4's exceptional performance on certain general and reasoning tasks~\cite{rein2023gpqa,hendrycks2021measuring}, it shows notably lower performance compared to Qwen2.5-14B with similar parameters, indicating the benchmark's difficulty.
\item\textbf{No single model excels across all categories, with different models showing distinct strengths.} For Service Fundamentals, Service with Context, User-Service Interaction, and Composite Tasks, the best-performing models are Qwen2.5-72B, GPT-4o, Claude 3.5-Sonnet, and Qwen2.5-72B, respectively. Notably, while Claude 3.5-Sonnet leads in User-Service Interaction, it performs worse than the lightweight Qwen2.5 Turbo in Service with Context tasks, demonstrating the varying capabilities of different models.
\end{itemize}
\begin{figure}[h]
\centering
\subfigure[Task-wise Score Correlations]{               
\includegraphics[width=3.5cm]{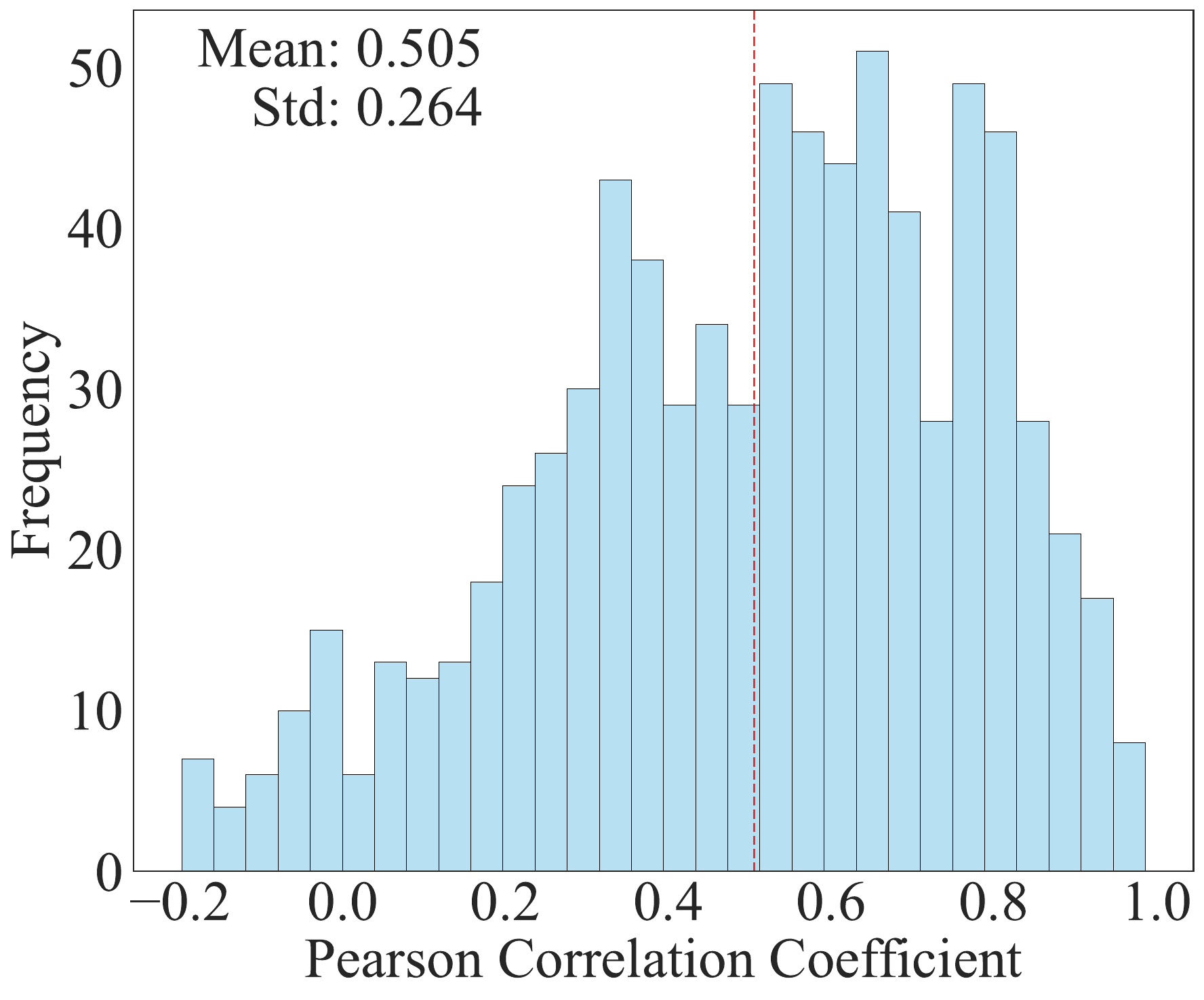}\label{fig:distribution}}
\hspace{0in}
\subfigure[Category-wise Correlations]{
\includegraphics[width=3.5cm]{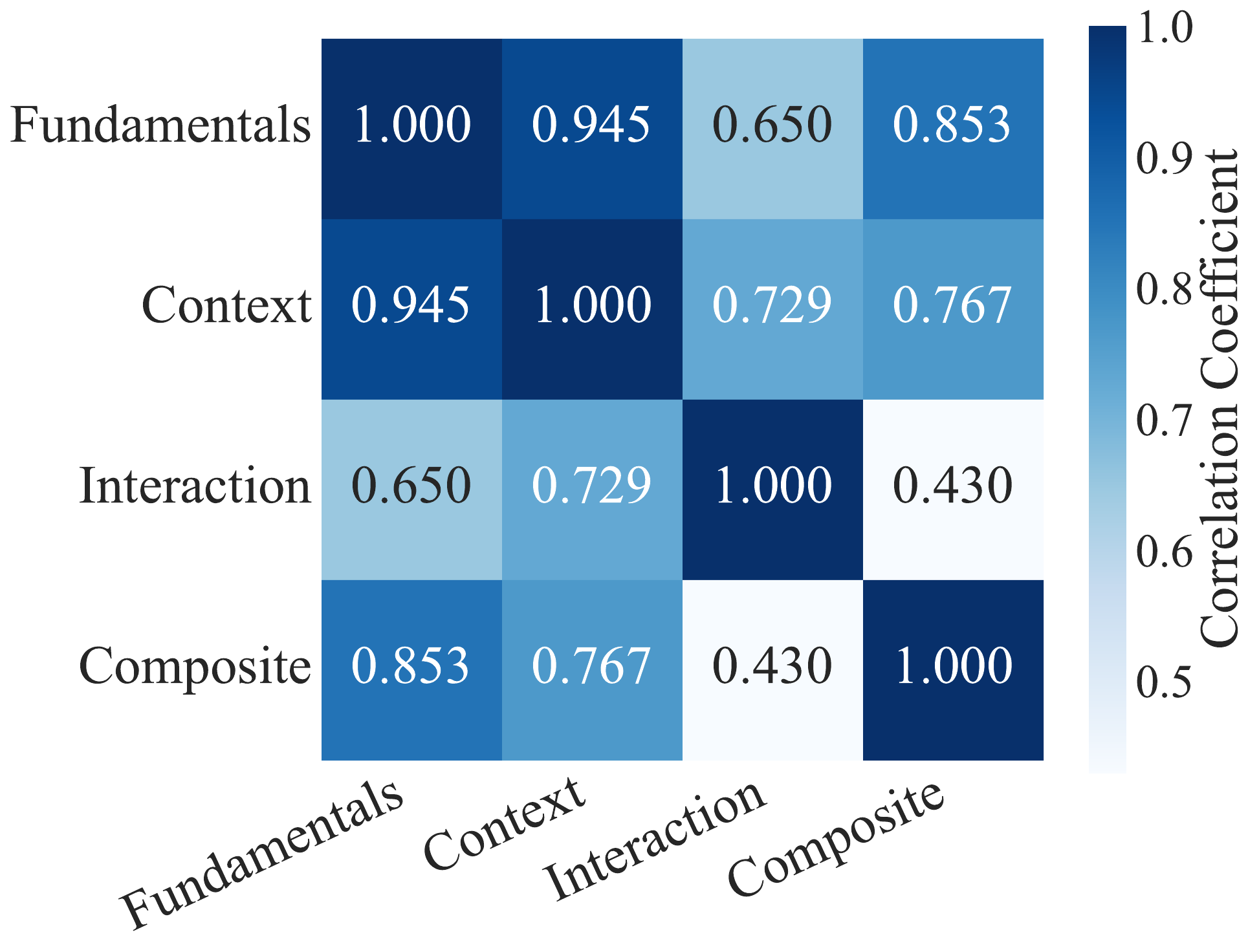}\label{fig:heatmap}}
\hspace{0in}
\vspace{-0.4cm}
\caption{Task and category-wise correlations on LocalEval.}\label{fig:correlation}
\vspace{-0.2cm}
\end{figure}

To better understand the relationships between tasks in the benchmark, we calculate the Pearson correlation between each pair of tasks, with the distribution of correlation coefficients shown in Figure~\ref{fig:distribution}. The results reveal overall positive correlation between tasks, with a mean correlation coefficient of 0.505 and a standard deviation of 0.264. This suggests that solving local life service understanding tasks requires common underlying knowledge and skills. We also analyze correlations between task categories, as shown in Figure~\ref{fig:heatmap}. Most task categories demonstrate significant positive correlations with each other. Notably, Composite Tasks show relatively lower correlations with other categories. This is likely due to their complexity, as models that possess the knowledge and capabilities to solve other tasks may still struggle with these composite problems. This observation supports the necessity of introducing agents to better address composite tasks.

\subsection{Instruction Tuning (RQ2)}
\begin{figure*}[]        
\includegraphics[width=16cm]{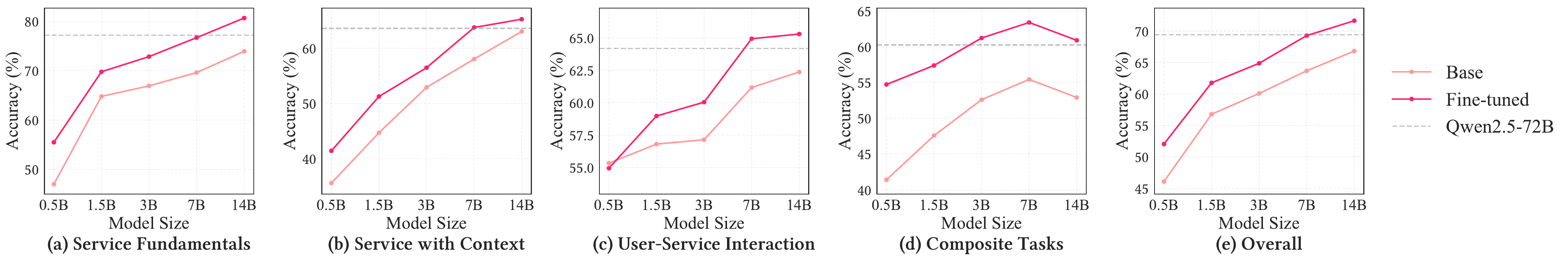}
\caption{Results of instruction tuning on Qwen2.5 series. Through fine-tuning on LocalInstruction, the performance of Qwen2.5-7B can match the performance of much larger Qwen2.5-72B.}
\label{fig:qwentuning}
\end{figure*}

\begin{figure*}[]        
\includegraphics[width=16cm]{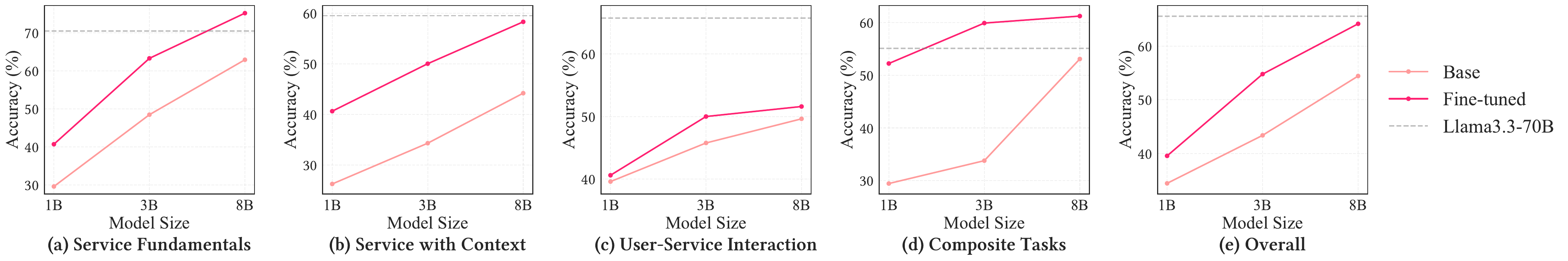}
\caption{Results of instruction tuning on Llama3 series. Through fine-tuning on LocalInstruction, the performance of Llama3.1-8B can match the performance of much larger Llama3.3-70B.}
\label{fig:llamatuning}
\end{figure*}

We fine-tune various LLMs using LocalInstruction, including five models from the Qwen2.5 Series (0.5B, 1.5B, 3B, 7B, 14B parameters) and three models from Llama (Llama3.2-1B, Llama3.2-3B, Llama3.1-8B). The experimental results are shown in Figure~\ref{fig:qwentuning} and~\ref{fig:llamatuning}. We observe significant performance improvements across all task categories for nearly all fine-tuned models. After our fine-tuning, the overall performance of Qwen2.5-7B approaches that of Qwen2.5-72B, while Llama3.1-8B's overall performance approaches that of Llama3.3-70B. Moreover, Qwen2.5-14B consistently outperforms Qwen2.5-72B on all task categories. These results demonstrate the effectiveness of our fine-tuning approach.
\begin{table}[]
\caption{Cross-city evaluation results (\%) of Qwen2.5-7B on the Service with Context category, showing performance when trained on data from Beijing, Shantou, Chongqing, and Urumqi, and tested across these cities.}
\vspace{-0.3cm}
\setlength{\tabcolsep}{0.4mm}
\begingroup\small
\resizebox{1\linewidth}{!}{
\begin{tabular}{c|cccccc}
\hline
\textbf{Test City}& \textbf{Base} & \textbf{Beijing} & \textbf{Shantou} & \textbf{Chongqing} & \textbf{Urumqi} & \textbf{All Data}  \\ \hline
\textbf{Beijing}  & 58.06  & 63.78            & 59.15            & 60.49              & 57.77           & 64.02                     \\
\textbf{Shantou}  & 42.31   & 43.83            & 48.33            & 45.03              & 44.34           & 48.65                    \\
\textbf{Chongqing} & 47.63 & 50.06            & 50.65            & 52.99              & 51.13           & 50.13                      \\
\textbf{Urumqi}  & 45.98   & 42.40            & 46.14            & 46.34              & 50.15           & 48.93                     \\ \hline
\end{tabular}}
\vspace{-0.1cm}
\endgroup

\label{tab:crosscontext}
\end{table}

\begin{table}[]
\caption{Cross-city evaluation results (\%) of Qwen2.5-7B on the User-Service Interaction category.} 
\vspace{-0.3cm}
\setlength{\tabcolsep}{0.4mm}
\begingroup\small
\resizebox{1\linewidth}{!}{
\begin{tabular}{c|cccccc}
\hline
\textbf{Test City} & \textbf{Base} & \textbf{Beijing} & \textbf{Shantou} & \textbf{Chongqing} & \textbf{Urumqi} & \textbf{All Data}  \\ \hline
\textbf{Beijing}  & 61.18    & 64.95            & 62.12            & 63.15              & 62.90           & 65.06                   \\
\textbf{Shantou} & 61.18  & 61.50            & 65.57            & 63.11              & 61.97           & 66.11                      \\
\textbf{Chongqing} & 63.22 & 64.60            & 64.04            & 66.35              & 64.35           & 66.40                     \\
\textbf{Urumqi}   & 56.76 & 62.53            & 57.43            & 59.12              & 64.06           & 64.95                      \\ \hline
\end{tabular}}
\endgroup
\label{tab:crossuser}
\end{table}
Local life services are highly location-dependent. Different cities have distinct geographical spaces and user lifestyles, requiring different knowledge to understand their local life services. In our main experiments, the benchmark and instruction tuning datasets primarily use merchant and user behavior data from Beijing. We attempt to build benchmarks and instruction tuning datasets for different cities to examine whether our instruction tuning method could enhance LLMs' understanding of local life services in different cities. Additionally, we analyze whether the capabilities of models trained on one city's data could transfer to another city. We conduct experiments on Qwen2.5-7B, selecting Shantou, Chongqing, and Urumqi as test cities due to their significant differences from Beijing in geographical features and user habits. When constructing training sets and benchmarks for other cities, we only change the selected cities while maintaining the same methodology and data volume for constructing training data and test questions. The results are shown in Table~\ref{tab:crosscontext} and Table~\ref{tab:crossuser}. Our findings include:
\begin{itemize}[leftmargin=*]
\item Our fine-tuning method is effective across all cities. Performance improvements are observed for all cities in both task categories when fine-tuning with data from the respective city.
\item Model capabilities generally demonstrate partial transfer between cities. In most cases, even when trained on another city's data, the model shows improved performance on the target city's tasks, although the improvement is not as substantial as when trained on the target city's own data. Training the model with data from all four cities, while specifying the current city during both training data construction and testing, achieves near-optimal results across both task categories and in all four models.
\item LLM capabilities struggle to transfer between cities with significant geographical and cultural differences. For the Service with Context task category, fine-tuning on Beijing data lead to decreased performance when testing on Urumqi, and vice versa. This suggests that training a universally effective service understanding model requires incorporating data from diverse cities to ensure the model comprehends different geographical environments and cultures.
\end{itemize}

To further validate our approach, we compared fine-tuning against various prompting strategies on Qwen2.5-7B, including role-playing, Chain-of-Thought (CoT), and few-shot learning. As shown in Table~\ref{tab:prompt_vs_finetune}, while these prompting techniques provide some performance gains (except for role-playing), they are significantly outperformed by our fine-tuning method. Moreover, few-shot learning shows diminishing returns after 10 shots and substantially increases inference costs, making it impractical for large-scale deployment. This analysis confirms that fine-tuning is a more effective and efficient strategy for domain adaptation in local life services.

\begin{table}[h!]
\centering
\caption{Performance comparison (\%) of different prompting strategies versus fine-tuning on Qwen2.5-7B. Fine-tuning consistently outperforms all prompting-based methods.}
\label{tab:prompt_vs_finetune}
\setlength{\tabcolsep}{0.4mm}
\resizebox{\linewidth}{!}{%
\begin{tabular}{lccccc}
\toprule
\textbf{Method} & \textbf{\begin{tabular}[c]{@{}c@{}}Service \\ Fundamentals\end{tabular}} & \textbf{\begin{tabular}[c]{@{}c@{}}Service with \\ Context\end{tabular}} & \textbf{\begin{tabular}[c]{@{}c@{}}User-Service \\ Interaction\end{tabular}} & \textbf{\begin{tabular}[c]{@{}c@{}}Composite \\ Tasks\end{tabular}} & \textbf{Overall} \\
\midrule
Base & 69.61 & 58.06 & 61.18 & 55.43 & 63.69 \\
\midrule
Role-playing & 70.48 & 57.39 & 60.03 & 52.23 & 63.37 \\
CoT & 71.91 & 57.22 & 61.29 & 58.08 & 64.73 \\
5-shot & 74.30 & 60.50 & 60.10 & 58.89 & 66.30 \\
10-shot & 75.62 & 60.22 & 61.09 & 59.22 & 67.08 \\
50-shot & 75.72 & 60.56 & 61.43 & 58.88 & 67.26 \\
\midrule
\textbf{Fine-tuned} & \textbf{76.68} & \textbf{63.78} & \textbf{64.95} & \textbf{63.88} & \textbf{69.29} \\
\bottomrule
\end{tabular}
}
\end{table}

\begin{figure}[]        
\includegraphics[width=7cm]{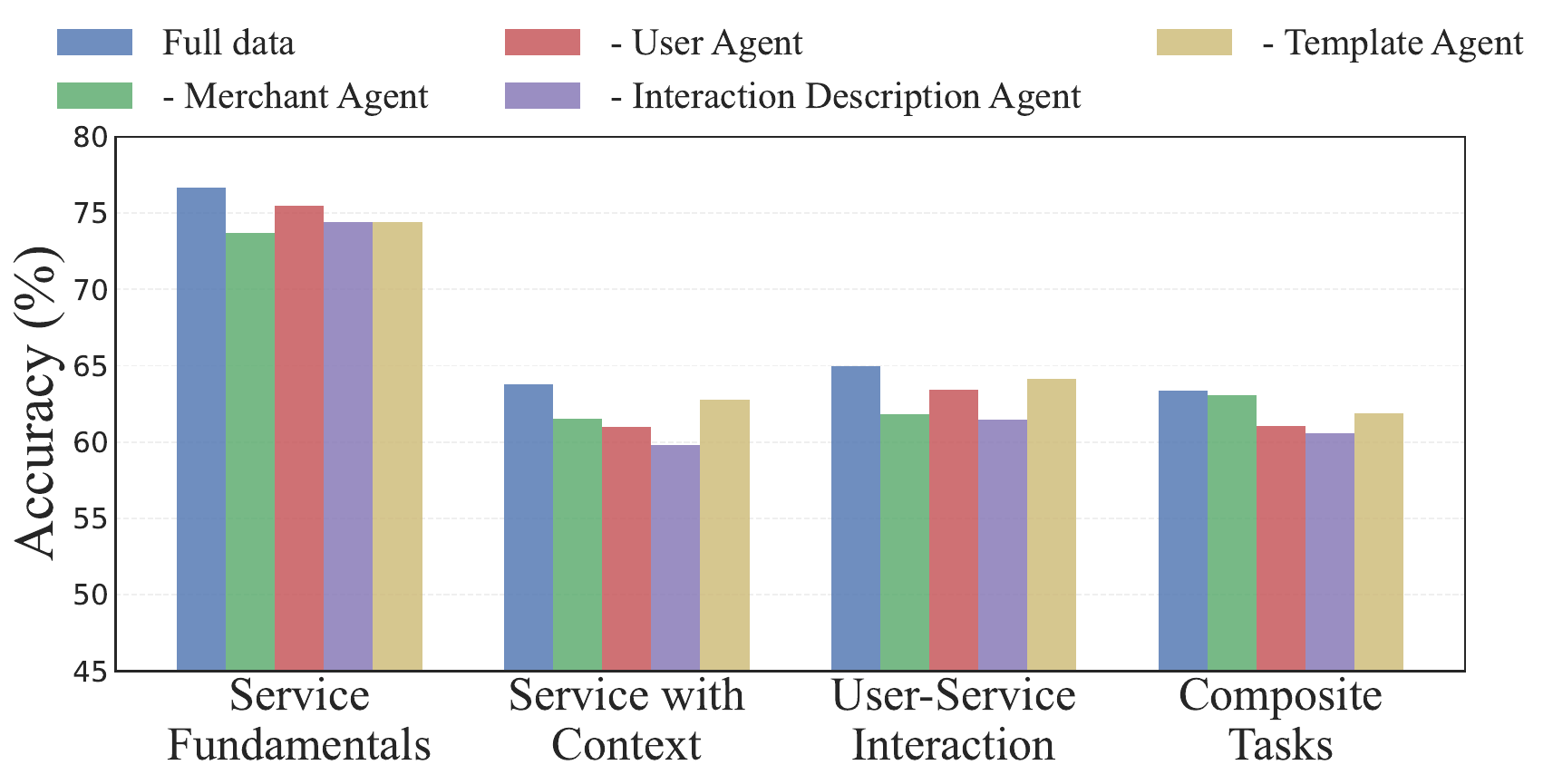}
\vspace{-0.3cm}
\caption{Results of ablation studies on Qwen2.5-7B.}
\vspace{-0.1cm}
\label{fig:ablation_qwen}
\end{figure}

\begin{figure}[]        
\includegraphics[width=7cm]{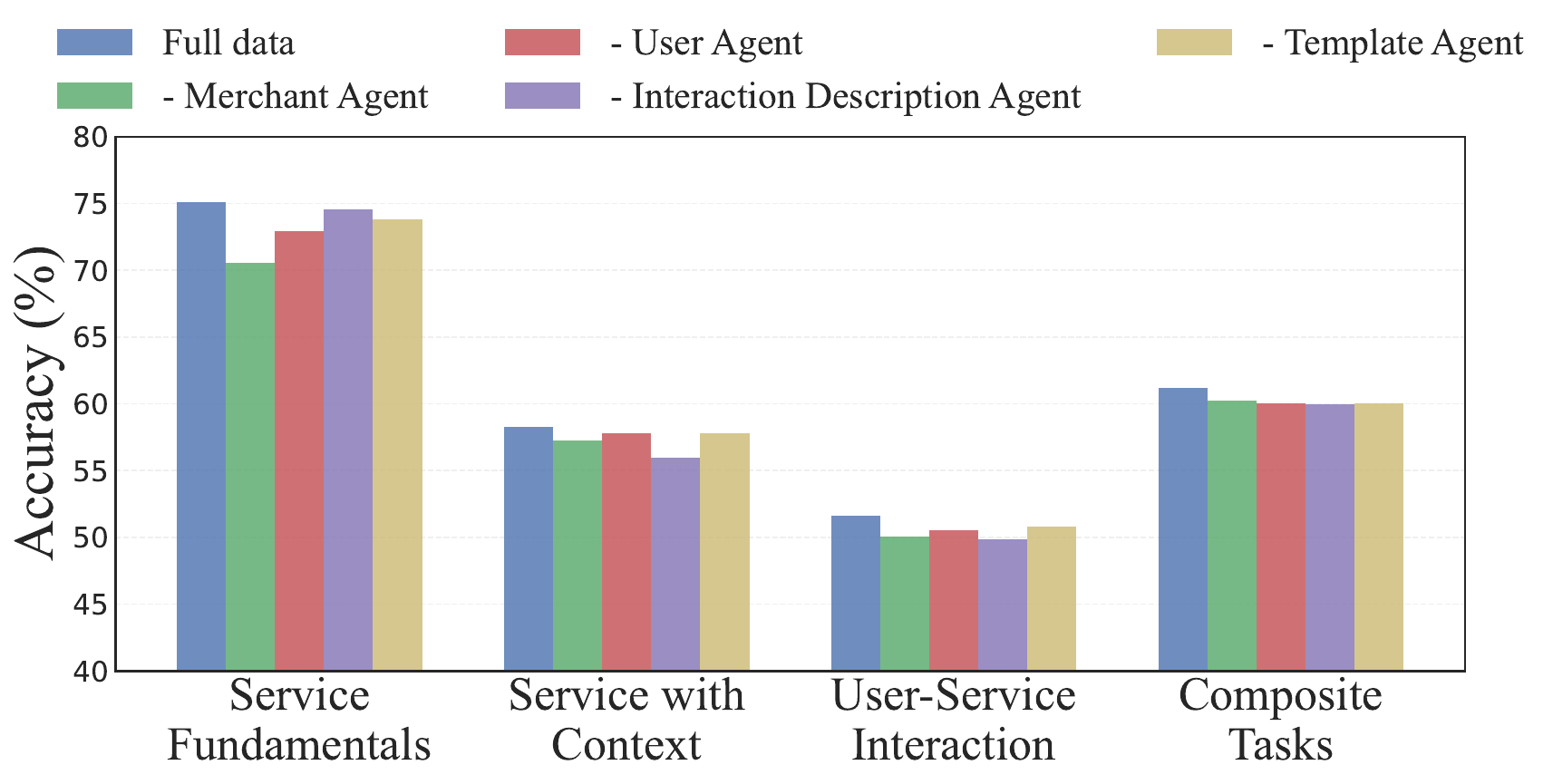}
\vspace{-0.3cm}
\caption{Results of ablation studies on Llama3.1-8B.}
\vspace{-0.4cm}
\label{fig:ablation_llama}
\end{figure}

\subsection{Ablation Study (RQ3)}
We conduct comprehensive ablation experiments by removing data generated by each type of agent and observing the impact on the performance of the fine-tuned model to verify whether data from each agent in LocalInstruction contributes to the final performance. We experiment with Qwen2.5-7B and Llama3.1-8B, with results shown in Figure~\ref{fig:ablation_qwen} and~\ref{fig:ablation_llama}. The results reveal the following observations:
\begin{itemize}[leftmargin=*]
\item Training data generated by each type of agent is effective across both models and all task categories. For both models and each task category, removing any type of training data leads to performance degradation, demonstrating the effectiveness of each component in our approach.
\item The Merchant Agent is most crucial for the Service Fundamentals task category. Across both datasets, removing the Merchant Agent has the largest negative impact on the Service Fundamentals task category, indicating that data generated by this agent contributes most significantly to improvements in this category. This shows that even though the Template Agent contains original merchant information, establishing logical connections through the LLM agent approach remains important.
\item The Interaction Description Agent is most critical for Service with Context and User-Service Interaction categories. Across both datasets, removing training data generated by the Interaction Description Agent has the most severe negative impact on the User-Service Interaction category, and its impact on Service with Context is also the largest among all data types. This demonstrates that data on user-merchant interactions within spatiotemporal contexts is crucial for LLMs to understand the relationships between services, users, and spatiotemporal factors.
\end{itemize}

\begin{table}[]
\small

\caption{Performance (\%)  of Qwen2.5-72B advanced by agentic workflows, and 7B fine-tuned with agent generated data.}
\label{tab:agent}
\vspace{-0.2cm}
\setlength{\tabcolsep}{0.4mm}
\begingroup\small
\resizebox{1\linewidth}{!}{
\begin{tabular}{cccc}
\hline
\textbf{Model}                        & \textbf{Recommend} & \textbf{Search} & \textbf{Review} \\ \hline
\textbf{Qwen2.5-72B with No Workflow}                  & 71.5                    & 57.2            & 48.0             \\
\textbf{Qwen2.5-72B+Agentic Workflow} & 77.0                    & 65.8            & 68.5             \\
\textbf{Qwen2.5-7B with No Workflow}                   & 68.5                    & 49.8            & 44.5             \\
\textbf{Qwen2.5-7B-Finetuned}         & 74.5                    & 57.5            & 59.5             \\ \hline
\end{tabular}}
\vspace{-0.5cm}
\endgroup
\end{table}
\subsection{Agentic Workflow (RQ4)}

To better address composite tasks that require multiple capabilities, we design agentic workflows to help large language models solve these problems more effectively. The data generated during problem-solving can then serve as instruction-tuning training data to improve model performance. In our experiments, we prompt Qwen2.5-72B solve problems following the workflow; the data generated by Qwen2.5-72B during problem-solving is used to train Qwen2.5-7B. The experimental results are shown in Table~\ref{tab:agent}. The results indicate that after following the workflow, Qwen2.5-72B's performance improves significantly, particularly in determining the most interesting reviews for users, where accuracy increases by 20.5\%. When the agent-generated data is used to train Qwen2.5-7B, it achieves substantial accuracy improvements across all three tasks. This demonstrates the effectiveness of designing agentic workflows for composite tasks, both as a problem-solving method and as a data generation process.

\section{Deployment and Applications}
\begin{table}[]
\small
\caption{LLM-generated merchant tags enhance the online recommendation system in Meituan.}
\vspace{-0.3cm}
\begin{tabular}{c|cccc}
\hline
\textbf{Category}      & PV & UV & OV & NCC \\ \hline
\textbf{Food}          & -0.09\%            & +0.00\%                   & +0.33\%                & +1.49\%                 \\
\textbf{Entertainment} & +1.75\%             & +1.89\%                   & +0.53\%                & +3.29\%                 \\
\textbf{Beauty}        & -0.59\%             & +0.25\%                   & +4.73\%             & +2.83\%                 \\
\textbf{Overall}       & +0.34\%             & +0.21\%                   & +0.27\%                & +0.15\%                 \\ \hline
\end{tabular}
\vspace{-0.1cm}
\label{tab:rec}
\end{table}

\begin{table}[]
\small
\caption{LLM-generated search query suggestions enhance the online search service in Meituan.}
\vspace{-0.3cm}
\setlength{\tabcolsep}{0.9mm}
\begin{tabular}{c|cccc}
\hline
Metric      & Session-CTR & UV-CTR & Query Views & Order Volume \\ \hline
Improv. & +0.48\%              & +0.44\%         & +0.33\%              & +0.53\%               \\ \hline
\end{tabular}
\vspace{-0.3cm}
\label{tab:search}
\end{table}
The fine-tuned large language models demonstrate enhanced understanding of Meituan's business operations and can deliver business value through practical deployment. Below, we present three deployment cases of our fine-tuned models in Meituan's recommendation, search, and review ranking scenarios. Note that our models possess general capabilities for local life service tasks; these three deployment cases utilize only a portion of the model's capabilities, leaving significant room for exploring other applications. Due to online latency constraints, our primary application method involves pre-computing tags or features for merchants and reviews, which are then used in online algorithms. The specific applications and experimental results are described below.

\textbf{Recommendation.} Users visit local life service platforms to fulfill their daily needs. Each merchant serves specific functions that satisfy particular user needs. For example, high-end steakhouses meet users' romantic dating needs, while Chinese restaurants fulfill group dining requirements. The original merchant data does not directly provide these important functional tags for understanding merchants. We use our fine-tuned model, trained on Meituan's business data, to generate merchant function tags based on merchant names, categories, and self-description texts. These tags become product features, and tag-based recall forms part of the online recommendation model's recall sources. We conduct a 7-day online A/B experiment on Meituan's homepage recommendation system to evaluate the performance improvement after adding model-generated function tags. The results are shown in Table~\ref{tab:rec}.

We present results for three representative categories and overall performance. In the table, PV refers to Page View, UV to Unique Visitors, OV to Order Volume, and NCC to New Customer Count. The Improvement indicates the performance increase after introducing LLM-based function tag recall compared to the baseline without this recall source. Most metrics show improvements, with significant increases in OV, which directly impacts platform revenue, across all categories. Notably, in the relatively long-tail beauty category, incorporating function tags for merchant modeling leads to a substantial 4.73\% increase in OV, demonstrating the high value of LLM-generated function tags.

\textbf{Search.} In search scenarios, suggesting search keywords based on partially entered content can enhance user experience. For example, if a user wants to buy \textit{Ketoconazole Ointment}, ideal keyword suggestions should appear after typing just the first few letters. However, meaningful search keywords are numerous, making manual definition labor-intensive. Therefore, we use the fine-tuned LLM to generate potential search queries from product descriptions, categories, and attributes to complement manually defined suggestion sets. In an online A/B Test conducted from July 19, 2024, to July 25, 2024, the experimental results in Table~\ref{tab:search} show consistent improvements across key engagement and conversion metrics. Users interact more frequently with the search suggestions, as reflected in the 0.48\% increase in Session-CTR and 0.44\% increase in UV-CTR. Most importantly, this enhanced search experience translates to business impact, with order volume increasing by 0.53\%.

\textbf{Review Ranking.} Ranking reviews displayed on product pages helps users find high-quality, relevant review content. Trustworthy and reference-worthy reviews should include: in-depth text content, actionable suggestions, natural language expression, credible and engaging language, non-promotional content, and non-AI-generated content. Traditional models struggle to identify these characteristics; we use our fine-tuned LLM to score reviews on these dimensions as review features. After incorporating these features into the ranking model, we observe significant improvements in review engagement and conversion. In an online A/B Test conducted
from January 18, 2025, to February 7, 2025, average review viewing duration increases by 1.42\%, average number of reviews viewed per user increases by 0.79\%, and the conversion rate of users who read reviews increases by 0.27\%.

In conclusion, experiments across these three scenarios demonstrate the high value of deploying our fine-tuned LLMs in actual applications on local life service platforms.

\section{Related Work}
\subsection{Large Language Models}
Large language models have witnessed rapid development. Since GPT-3~\cite{brown2020language} demonstrated strong few-shot learning capabilities through scaling, more powerful models like Claude-3.5 Sonnet~\cite{anthropic2024introducing} and GPT-4~\cite{achiam2023gpt} have shown remarkable performance in reasoning and knowledge integration. These models excel in text analysis~\cite{ziems2024can,lan2024stance,lan2024depression}, mathematical reasoning~\cite{lewkowycz2022solving,xu2025towards,hao2025rl,hao2025llm}, and scientific analysis~\cite{hao2024hlm}. Due to these diverse capabilities, LLMs have been successfully applied in various domains, including social sciences~\cite{gao2023s3,gao2024large,li2023econagent,piao2025agentsociety,wang2024survey}, natural sciences~\cite{li2025materials}, graph learning~\cite{li2023survey,li2023gslb,zhang2024cut}, software development~\cite{zhang2023survey,qian2024chatdev}, healthcare~\cite{li2023chatdoctor,thirunavukarasu2023large,al2023transforming}, urban science~\cite{feng2024citygpt, feng2025survey, ding2024understanding}, and education~\cite{jeon2023large,kasneci2023chatgpt}. While local life services represent a crucial domain impacting billions of users' daily lives, systematic research on applying LLMs to understand local life services and help users better interact with local life services remains unexplored. Our work bridges this research gap by comprehensively evaluating and improving LLMs' capabilities in this domain.

\subsection{Evaluation Benchmark}
Large language models (LLMs) have emerged as general-purpose models, prompting extensive efforts to evaluate their capabilities through diverse benchmarks. These evaluations cover fundamental language understanding~\cite{wang2018glue,wang2019superglue}, reasoning abilities~\cite{srivastava2022beyond,hendrycks2021measuring}, domain knowledge~\cite{kasai2023evaluating,guha2024legalbench,jin2024shopping,feng2024citybench}, and specialized skills~\cite{chen2021evaluating,talmor2018commonsenseqa}. Furthermore, researchers have developed benchmarks for assessing specific cognitive capabilities like commonsense understanding~\cite{bisk2020piqa}, moral judgment~\cite{ji2024moralbench}, and in-depth thinking~\cite{zeng2023challenge}. However, despite the significant economic and social impact of local life services in modern society, there has been no systematic evaluation of LLMs' capabilities in this crucial domain. Our work addresses this gap by introducing LocalEval, the first comprehensive benchmark for assessing LLMs' abilities in local life services.
\subsection{LLM-Based Agents}
Large language models (LLMs) possess broad general capabilities. When applying LLMs to downstream tasks, building LLM agents proves beneficial, as the agent paradigm provides optimized problem-solving workflows while enabling external memory and tool integration~\cite{xi2025rise,wang2024survey}. LLM agents have achieved excellent results across various domains, including code generation~\cite{qian2024chatdev,qian2023communicative}, text analysis~\cite{lan2024stance}, recommendation systems~\cite{hou2024large,zhu2023large}, embodied intelligence~\cite{zhang2024smartagent}, social simulation~\cite{li2024econagent}, and urban science~\cite{feng2025survey,feng2025agentmove}. Some recent work has also utilized agents for generating instruction tuning data, where agent workflows ensure high-quality data generation while the use of diverse agents as generation seeds guarantee data variety~\cite{bi2023oceangpt,ou2024synatra,zhou2024star,ge2024scaling}. Inspired by these two lines of work, in our approach, we leverage agents to generate instruction-tuning data for fine-tuning LLMs, while also designing agentic workflows to help LLMs better solve complex problems.

\section{Conclusion}
In this paper, we introduce a framework for evaluating and enhancing LLMs in local life service platforms. We develop a comprehensive benchmark with over 40 tasks, and propose a multi-agent based instruction tuning approach that enables 7B parameter models to achieve competitive performance with larger 70B parameter models. Through extensive experiments and real-world deployments in Meituan's recommendation, search, and review ranking scenarios, we demonstrate the effectiveness of our framework in a wide range of tasks. Our work provides a foundation for developing and evaluating domain-specific LLMs for local life service platforms.
\begin{acks}
This work was supported in part by the National Key Research and Development Program of China under grant 2024YFC3307603, in part by the China Postdoctoral Science Foundation under grant 2024M761670 and GZB20240384, in part by the Tsinghua University Shuimu Scholar Program under grant 2023SM235. This work was also supported by Meituan.
\end{acks}
\clearpage
\bibliographystyle{ACM-Reference-Format}
\bibliography{reference}

\appendix
\section{Appendix}
\subsection{LocalEval Benchmark Task Details}
\label{sec:appendix}

This appendix provides a comprehensive list of all tasks included in the LocalEval benchmark. The benchmark consists of 41 tasks organized into four main categories, designed to systematically evaluate LLMs' capabilities in understanding and reasoning about local life services.

\subsubsection{Service Fundamentals}

This category evaluates the model's understanding of basic service properties including categories, attributes, values, and their relationships.

\begin{itemize}
    [leftmargin=*]
    \setlength{\itemsep}{0pt}
    \setlength{\parsep}{0pt}
    \setlength{\parskip}{0pt}
    \item \textbf{Category Prediction}: Given merchant name and related products, determine the merchant's category.
    \item \textbf{Attribute Mining}: Given merchant name and related products, select applicable attributes.
    \item \textbf{Attribute Value Extraction}: Given merchant description and target attribute dimension, extract corresponding values.
    \item \textbf{Multi-level Category Prediction}: Given merchant name, determine complete category path from top-level to finest granularity.
    \item \textbf{Category-based Merchant Selection}: Given merchant category, select most likely merchant names.
    \item \textbf{Attribute-based Category Selection}: Given merchant attribute, select applicable categories.
    \item \textbf{Same-category Judgment}: Determine if two merchant names belong to the same category.
    \item \textbf{Same-category Selection}: Given a merchant name, select merchants from the same category.
    \item \textbf{Attribute Value Reasonableness}: Given merchant description and attribute dimension, judge value reasonableness.
    \item \textbf{Attribute Value Identification}: Given merchant description and attribute value, identify the described attribute.
    \item \textbf{Attribute Value Synonym Detection}: Determine if two attribute values express the same meaning.
    \item \textbf{Attribute Value Containment}: Determine if containment relationship exists between two attribute values.
    \item \textbf{Attribute Compatibility}: Determine if two attributes/values can describe the same merchant.
    \item \textbf{Mathematical Operations}: Given basic quantities, perform mathematical calculations.
    \item \textbf{Function Tag Prediction}: Given merchant description, predict function tags (e.g., suitable for family outing).
    \item \textbf{Brand Positioning}: Given two brands, determine which is more premium.
    \item \textbf{Brand Similarity}: Given a brand, select the most similar brand.
    \item \textbf{Category Complementarity}: Given a category, select complementary categories.
\end{itemize}

\subsubsection{Service with Context}

This category assesses understanding of spatiotemporal factors and their impact on local life services.

\begin{itemize}
    [leftmargin=*]
    \setlength{\itemsep}{0pt}
    \setlength{\parsep}{0pt}
    \setlength{\parskip}{0pt}
    \item \textbf{Weather Impact (Qualitative)}: Given merchant description, qualitatively judge weather impact on consumption.
    \item \textbf{Weather Impact (Quantitative)}: Given merchant description, quantitatively predict weather impact on consumption.
    \item \textbf{Seasonal Impact (Qualitative)}: Given merchant description, qualitatively judge seasonal impact on consumption.
    \item \textbf{Seasonal Impact (Quantitative)}: Given merchant description, quantitatively predict seasonal impact on consumption.
    \item \textbf{Nearest Merchant Selection}: Given merchant name and address, select nearest other merchants.
    \item \textbf{Distance Estimation}: Given two merchant addresses, estimate distance between them.
    \item \textbf{Administrative Division}: Given merchant name and landmark, select administrative district.
    \item \textbf{Business District Identification}: Given merchant name, select business district location.
    \item \textbf{Operating Hours Prediction}: Given merchant name, predict most likely operating hours.
    \item \textbf{Peak Hours Prediction}: Given merchant description, select daily consumption peak periods.
\end{itemize}

\subsubsection{User-Service Interaction}

This category evaluates understanding of user perspectives and preferences regarding local life services.

\begin{itemize}
    [leftmargin=*]
    \setlength{\itemsep}{0pt}
    \setlength{\parsep}{0pt}
    \setlength{\parskip}{0pt}
    \item \textbf{Target Group Identification}: Given service type, select suitable consumer groups.
    \item \textbf{User Preference Prediction}: Given user profile, predict most likely service consumption.
    \item \textbf{Review Information Points}: Count the number of informative points in a review.
    \item \textbf{Review Guidance Value}: Determine if review provides actionable suggestions.
    \item \textbf{Review Colloquialism}: Assess if review uses colloquial expressions.
    \item \textbf{Review Real Examples}: Determine if review includes real-world examples.
    \item \textbf{Review Language Appeal}: Assess if review language is engaging and persuasive.
    \item \textbf{Non-marketing Content}: Determine if review is free from promotional content.
    \item \textbf{Human-written Content}: Determine if review is not AI-generated or padded.
    \item \textbf{Overall Review Usefulness}: Judge overall usefulness of the review.
\end{itemize}

\subsubsection{Composite Tasks}

This category includes complex tasks requiring integration of multiple capabilities, corresponding to real-world application scenarios.

\begin{itemize}
    [leftmargin=*]
    \setlength{\itemsep}{0pt}
    \setlength{\parsep}{0pt}
    \setlength{\parskip}{0pt}
    \item \textbf{Recommendation}: Predict user consumption based on prior behavior sequences, user profiles, and spatiotemporal context.
    \item \textbf{Search}: Predict user clicks given ambiguous search queries, user profiles, and spatiotemporal context.
    \item \textbf{Content Marketing}: Identify reviews of most interest to users based on their profiles.
\end{itemize}

\subsection{Comparison with Alternative Instruction Synthesis Approaches}

To validate the effectiveness of our multi-agent-based instruction synthesis method, we compare LocalInstruction with two alternative data generation approaches:

\begin{itemize}
    [leftmargin=*]
    \setlength{\itemsep}{0pt}
    \setlength{\parsep}{0pt}
    \setlength{\parskip}{0pt}
    \item \textbf{DataOnly}: Uses only necessary templates to organize raw data without any LLM-based processing or logical connection establishment.
    \item \textbf{SelfInstruct}: Employs a single LLM to directly transform data into instruction-output format without specialized agents.
\end{itemize}

All methods generate the same number of training examples and follow identical training strategies as described in the main paper. We fine-tune Qwen2.5-7B using data from each approach and evaluate on LocalEval benchmark.

\begin{table}[h]
\centering
\caption{Performance (\%) comparison of different instruction synthesis methods on Qwen2.5-7B.}
\vspace{-0.3cm}
\setlength{\tabcolsep}{0.4mm}
\label{tab:instruction_comparison}
\begingroup\small
\resizebox{1\linewidth}{!}{
\begin{tabular}{lccccc}
\toprule
\textbf{Method} & \textbf{Service} & \textbf{Service with} & \textbf{User-Service} & \textbf{Composite} & \textbf{Overall} \\
& \textbf{Fundamentals} & \textbf{Context} & \textbf{Interaction} & \textbf{Tasks} & \\
\midrule
DataOnly & 74.38 & 60.39 & 61.97 & 58.96 & 66.78 \\
SelfInstruct & 74.21 & 61.09 & 62.52 & 60.98 & 67.17 \\
Our Method & \textbf{76.68} & \textbf{63.78} & \textbf{64.95} & \textbf{63.88} & \textbf{69.29} \\
\midrule
Improvement & +2.30 & +2.69 & +2.43 & +2.90 & +2.12 \\
\bottomrule
\end{tabular}}
\vspace{-0.2cm}
\endgroup
\end{table}

The results demonstrate that our multi-agent approach consistently outperforms both baseline methods across all task categories. The DataOnly approach, which lacks LLM-based logical connections, shows the lowest performance. While SelfInstruct improves upon DataOnly by leveraging LLM capabilities, it still falls short of our method. Our approach achieves an overall improvement of 2.12\% over the stronger SelfInstruct baseline, with particularly notable gains in Composite Tasks (+2.90\%), validating the effectiveness of specialized agents in generating high-quality instruction data.
\end{document}